\documentclass[10pt,twocolumn,letterpaper]{article}

\pdfoutput=1

\usepackage{cvpr}
\usepackage{times}
\usepackage{epsfig}
\usepackage{graphicx}
\usepackage{epstopdf}
\usepackage{amsmath}
\usepackage{amssymb}
\usepackage{caption}
\usepackage{booktabs}
\usepackage{dblfloatfix} 
\usepackage[T1]{fontenc}
\usepackage{kotex}

\usepackage[breaklinks=true,bookmarks=false]{hyperref}

\cvprfinalcopy 

\def\cvprPaperID{6930} 

\newenvironment{myindentpar}[1]%
  {\begin{list}{}%
          {\setlength{\leftmargin}{#1}}%
          \item[]%
  }
  {\end{list}}
  
\begin{document}

\title{Image-to-Image Translation via Group-wise Deep Whitening-and-Coloring Transformation}

\author{$\text{Wonwoong Cho}^{1}$
\and
$\text{Sungha Choi}^{1,2}$
\and
$\text{David Keetae Park}^{1}$
\and
$\text{Inkyu Shin}^{3}$
\and
$\text{Jaegul Choo}^{1}$
\vspace{0.1em}\and
$^{1}\text{Korea University}$
\and
$^{2}\text{LG Electronics}$
\and
$^{3}\text{Hanyang University}$
}

\maketitle

\begin{abstract}
   Recently, unsupervised exemplar-based image-to-image translation, conditioned on a given exemplar without the paired data, has accomplished substantial advancements. In order to transfer the information from an exemplar to an input image, existing methods often use a normalization technique, e.g., adaptive instance normalization, that controls the channel-wise statistics of an input activation map at a particular layer, such as the mean and the variance. Meanwhile, style transfer approaches similar task to image translation by nature, demonstrated superior performance by using the higher-order statistics such as covariance among channels in representing a style. In detail, it works via whitening (given a zero-mean input feature, transforming its covariance matrix into the identity). followed by coloring (changing the covariance matrix of the whitened feature to those of the style feature). However, applying this approach in image translation is computationally intensive and error-prone due to the expensive time complexity and its non-trivial backpropagation. In response, this paper proposes an end-to-end approach tailored for image translation that efficiently approximates this transformation with our novel regularization methods. We further extend our approach to a group-wise form for memory and time efficiency as well as image quality. Extensive qualitative and quantitative experiments demonstrate that our proposed method is fast, both in training and inference, and highly effective in reflecting the style of an exemplar. Finally, our code is available at \url{https://github.com/WonwoongCho/GDWCT}.
\end{abstract}

\section{Introduction}\label{intro}

Since the introduction of image-to-image translation~\cite{8100115}, in short, image translation, it has gained significant attention from relevant fields and constantly evolved propelled by the seminal generative adversarial networks~\cite{goodfellow2014generative}. The primary goal of image translation~\cite{8100115,Zhu_2017} is to convert particular attributes of an input image in an original domain to a target one, while maintaining other semantics. Early models for image translation required training data as paired images of an input and its corresponding output images, allowing a direct supervision. CycleGAN~\cite{Zhu_2017} successfully extends it toward unsupervised image translation~\cite{liu2017unsupervised,Chang:2018:PAS,StarGAN2018,Zhu_2017} by proposing the cycle consistency loss, which allows the model to learn the distinctive semantic difference between the collections of two image domains and translate the corresponding style without a direct pair-wise supervision. 

Nonetheless, CycleGAN is still unimodal in that it can only generate a single output for a single input. Instead, image translation should be capable of generating multiple possible outputs even for a single given input, e.g., numerous possible gender-translated outputs of a single facial image. Subsequently, two notable methods, DRIT~\cite{Lee_2018_ECCV} and MUNIT~\cite{Huang_2018_ECCV}, have been proposed to address the multimodal nature of unsupervised image translation. They demonstrate that a slew of potential outputs could be generated given a single input image, based on either a random sampling process in the midst of translation or utilizing an additional, exemplar image for a detailed guidance toward a desired style.

They both have two separate encoders corresponding to the content image (an input) and style image (an exemplar), and combine the content feature and style feature together to produce the final output. DRIT concatenates the encoded content and style feature vectors, while MUNIT exploits the adaptive instance normalization (AdaIN), a method first introduced in the context of style transfer. AdaIN matches two channel-wise statistics, the mean and variance, of the encoded content feature with the style feature, which is proven to perform well in image translation.     

However, we hypothesize that matching only these two statistics may not reflect the target style well enough, ending up with the sub-optimal quality of image outputs on numerous occasions, as we confirm through our experiments in Section~\ref{experiment}. 
That is, the interaction effects among variables, represented as the Gram matrix~\cite{gatys2016image} or the covariance matrix~\cite{li2017diversified}, can convey critical information of the style, which is agreed by extensive studies~\cite{gatys2016image,gatys2015texture,li2017diversified}. In response, to fully utilize the style information of an exemplar, we propose a novel method that takes into account such interaction effects among feature channels, in the context of image translation.

Our model is mainly motivated by whitening-and-coloring transformation (WCT)~\cite{li2017universal}, which utilizes the pair-wise feature covariances, in addition to the mean and the variance of each single feature, to encode the style of an image. To elaborate, whitening refers to the normalization process to make every covariance term (between a pair of variables) as well as every variance term (within each single variable) as a unit value, with given an input whose each single variable is zero-meaned. This plays a role in removing (or neutralizing) the style. On the other hand, coloring indicates the procedure of matching the covariance of the style to that of the content feature, which imposes the intended style into an neutralized input image. 

The problem when applying WCT in image translation is that its time complexity is as expensive as $O(n^3)$ 
where ${n}$ is the number of channels of a given activation map. Furthermore, computing the backpropagation with respect to singular value decomposition involved in WCT is non-trivial~\cite{Wei_2018_ECCV,ionescu2015matrix}. To address these issues, we propose a novel deep whitening-and-coloring transformation that flexibly approximates the existing WCT based on deep neural networks. We further extend our method into group-wise deep whitening-and-coloring transformation (GDWCT), which does not only reduce the number of parameters and the training time but also boosts the generated image quality~\cite{Wu_2018_ECCV,huang2018decorrelated}.

The main contribution of this paper includes:
\begin{myindentpar}{0.3cm}
\noindent$\bullet$ We present the novel deep whitening-and-coloring approach that allows an end-to-end training in image translation for conveying profound style semantics.  \\[0.3em]
$\bullet$ We also propose the group-wise deep whitening-and-coloring algorithm to further increase the computational efficiency through a simple forward propagation, which achieves highly competitive image quality. \\[0.3em]
$\bullet$ We demonstrate the effectiveness of our method via extensive quantitative and qualitative experiments, compared to state-of-the-art methods.
\end{myindentpar}

\section{Related Work}
\paragraph{Image-to-image translation.} Image-to-image translation aims at converting an input image to another image with a target attribute. Many of its applications exist, e.g., colorization~\cite{zhang2016colorful,deshpande2017learning,bahng2018coloring,yoo2019coloring}, super-resolution~\cite{dong2014learning,ledig2017photo}, and domain adaptation~\cite{hoffman18a,li2018learning}.


A slew of studies have been conducted in an unsupervised setting of image translation~\cite{Zhu_2017, kim2017learning, liu2017unsupervised}. 
StarGAN~\cite{StarGAN2018} proposes a single unified model which can handle unsupervised image translation among multiple different domains. 

Several studies~\cite{ghosh2018multi,NIPS2017_6650} focus on the limitation of earlier work in which they produce a single output given an input without consideration that diverse images can be generated within the same target domain. However, they are not without limitations, either by generating a limited number of outputs~\cite{ghosh2018multi} or requiring paired images~\cite{NIPS2017_6650}.

Recently proposed approaches~\cite{Huang_2018_ECCV, Lee_2018_ECCV} are capable of generating multimodal outputs in an unsupervised manner. They work mainly based on the assumption that a latent image space could be separated into a domain-specific style space a domain-invariant content spaces. Following the precedents, we also adopt the separate encoders to extract out each of the content and style features.

\vspace*{-0.25cm}
\paragraph{Style transfer.} Gatys et al.~\cite{gatys2015texture,gatys2016image} show that the pairwise feature interactions obtained from the Gram matrix or the covariance matrix of deep neural networks successfully capture the image style. It is used for transferring the style information from a style image to a content image by matching the statistics of the style feature with those of the content. However, they require a time-consuming, iterative optimization process during the inference time involving multiple forward and backward passes to obtain a final result. To address the limitation, alternative methods~\cite{ulyanov2016texture, chen2017stylebank, johnson2016perceptual} have achieved a superior time efficiency through feed-forward networks approximating an optimal result of the iterative methods. 

However, these models are incapable of transferring an unseen style from an arbitrary image. To alleviate the limitation, several approaches enable an unseen, arbitrary neural style transfer~\cite{huang2017arbitrary,li2017universal,li2018closed}. AdaIN~\cite{huang2017arbitrary} directly computes the affine parameters from the style feature and aligns the mean and variance of the content feature with those of the style feature. WCT~\cite{li2017universal} encodes the style as the feature covariance matrix, so that it effectively captures the rich style representation. Recently, a new approach~\cite{li2018learning} have approximated the whitening-and-coloring transformation as a one-time transformation through a single transformation matrix. Even though the idea of learning the transformation is similar to ours, the proposed networks are incapable of transferring the semantic style information, such as the translation between the cat and the dog because the existing approach can transfer only the general style, such as the color and the texture. Moreover, its settings of approximating the transformation is less rigorous than ours due to the lack of the regularization for ensuring the whitening-and-coloring transformation.

\begin{figure*}
\vspace*{-0.9cm}
  \includegraphics[width=\linewidth]{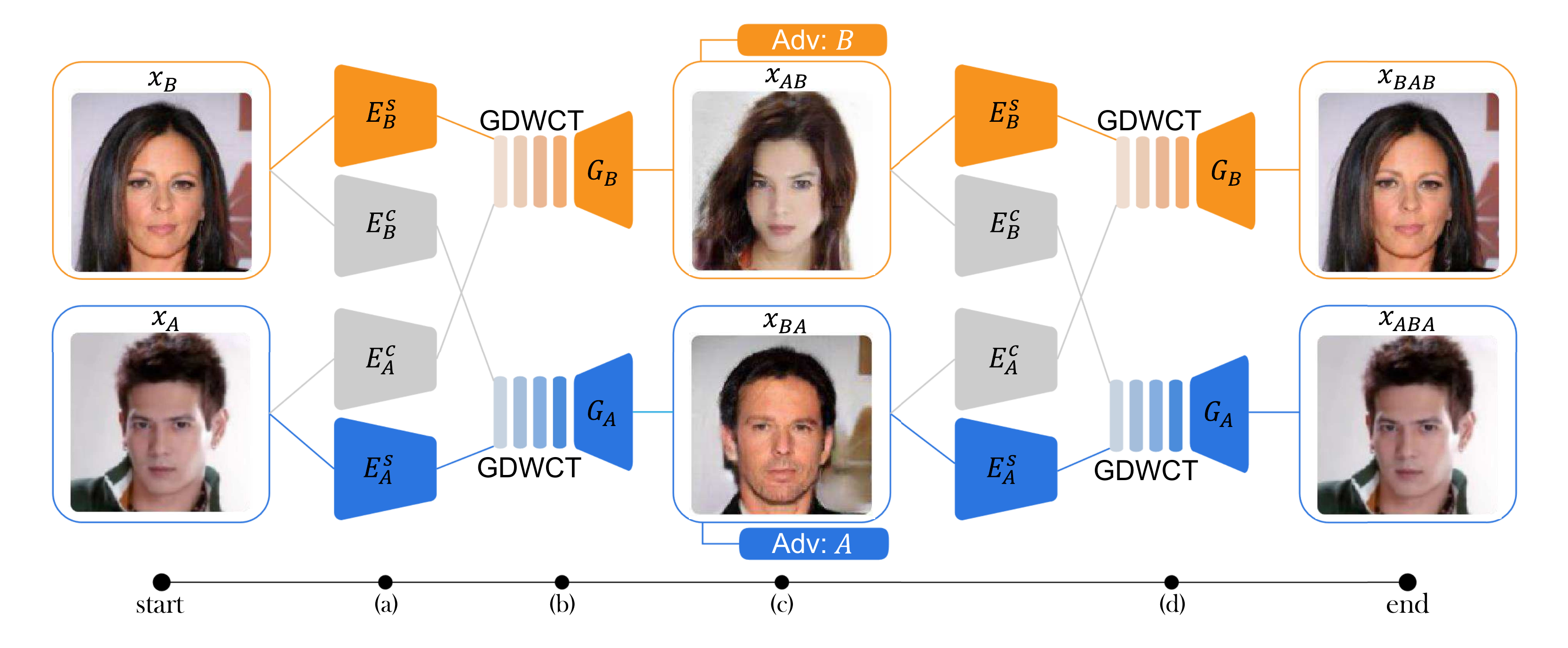}
  \vspace*{-0.9cm}
  \caption{Overview of our model. (a) To translate from ${\mathcal{A}\rightarrow{\mathcal{B}}}$, we first extract the content feature ${c_\mathcal{A}}$ from the image ${x_\mathcal{A}}$ (i.e., ${c_\mathcal{A}=E_\mathcal{A}^c(x_\mathcal{A})}$) and the style feature ${s_\mathcal{B}}$ from the image ${x_\mathcal{B}}$ (i.e., ${s_\mathcal{B}=E_\mathcal{B}^s(x_\mathcal{B})}$). (b) The obtained features are combined in our GDWCT module while forwarded through the generator ${G_\mathcal{B}}$. (c) The discriminator ${D_\mathcal{B}}$ classifies whether the input  ${x_{\mathcal{A}\mathcal{B}}}$ is a real image of the domain ${\mathcal{B}}$ or not. (d) Similar to the procedures from (a) to (c), the generator ${G_\mathcal{B}}$ generates the reconstructed image ${x_{\mathcal{B}\mathcal{A}\mathcal{B}}}$ by combining the content feature ${c_{\mathcal{B}\mathcal{A}}}$ and the style feature ${s_{\mathcal{A}\mathcal{B}}}$.}
  \label{overview_fig}
  \vspace*{-0.3cm}
\end{figure*}

\section{Proposed Method}
This section describes our proposed model in detail, by first giving a model overview and by explaining the our proposed loss functions.

\subsection{Model Overview}\label{model}
Let ${x_\mathcal{A}}\in\mathcal{X}_\mathcal{A}$ and ${x_\mathcal{B}}\in\mathcal{X}_\mathcal{B}$ denote images from two different image domains, $\mathcal{X}_\mathcal{A}$, $\mathcal{X}_\mathcal{B}$, respectively. Inspired by MUNIT~\cite{Huang_2018_ECCV} and DRIT~\cite{Lee_2018_ECCV}, we assume that the image ${x}$ can be decomposed into the domain-invariant content space ${\mathcal{C}}$ and the domain-specific style spaces \{${\mathcal{S}_\mathcal{A}}$, ${\mathcal{S}_\mathcal{B}}$\}, i.e.,
\begin{align*}
    \{c_\mathcal{A},s_\mathcal{A}\}&=\{E_\mathcal{A}^c(x_\mathcal{A}),E_\mathcal{A}^s(x_\mathcal{A})\}\qquad c_\mathcal{A} \in\mathcal{C},s_\mathcal{A} \in\mathcal{S}_\mathcal{A} \\[0.5em]
    \{c_\mathcal{B},s_\mathcal{B}\}&=\{E_\mathcal{B}^c(x_\mathcal{B}),E_\mathcal{B}^s(x_\mathcal{B})\}\qquad c_\mathcal{B} \in\mathcal{C},s_\mathcal{B} \in\mathcal{S}_\mathcal{B},
\end{align*}
where \{${E_\mathcal{A}^c}$, ${{E_\mathcal{B}^c}}$\} and \{${E_\mathcal{A}^s}$, ${E_\mathcal{B}^s}$\} are the content and style encoders for each domain, respectively. Our objective is to generate the translated image by optimizing the functions \{${f_{\mathcal{A}\rightarrow{\mathcal{B}}}}$, ${f_{\mathcal{B}\rightarrow{\mathcal{A}}}}$\} of which ${f_{\mathcal{A}\rightarrow{\mathcal{B}}}}$ maps the data point ${x_\mathcal{A}}$ in the original domain ${\mathcal{X}_\mathcal{A}}$ to the point ${x_{\mathcal{A}\rightarrow{\mathcal{B}}}}$ in the target domain ${\mathcal{X}_\mathcal{B}}$, reflecting a given reference ${x_\mathcal{B}}$, i.e., 
\begin{align*}
    x_{\mathcal{A}\rightarrow{\mathcal{B}}}&=f_{\mathcal{A}\rightarrow{\mathcal{B}}}(x_\mathcal{A},x_\mathcal{B})=G_\mathcal{B}(E_\mathcal{A}^c(x_\mathcal{A}),E_\mathcal{B}^s(x_\mathcal{B}))\\[0.5em]
    x_{\mathcal{B}\rightarrow{\mathcal{A}}}&=f_{\mathcal{B}\rightarrow{\mathcal{A}}}(x_\mathcal{B},x_\mathcal{A})=G_\mathcal{A}(E_\mathcal{B}^c(x_\mathcal{B}),E_\mathcal{A}^s(x_\mathcal{A})),
\end{align*}
where \{${G_\mathcal{A},G_\mathcal{B}}$\} are the generators for the corresponding domains. 

As illustrated in Fig.~\ref{overview_fig}, the group-wise deep whitening-and-coloring transformation (GDWCT), plays a main role in applying the style feature ${s}$ to the content feature ${c}$ inside the generator ${G}$. Concretely, GDWCT takes the content feature ${c_\mathcal{A}}$, the matrix for coloring transformation ${s_\mathcal{B}^\text{CT}}$, and the mean of the style ${s_\mathcal{B}^{\,\mbox{\footnotesize $\mu$}}}$ as input and conduct a translation of ${c_\mathcal{A}}$ to ${c_{\mathcal{A}\rightarrow{\mathcal{B}}}}$, formulated as
\begin{align*}
    c_{\mathcal{A}\rightarrow{\mathcal{B}}}=\text{GDWCT}(c_\mathcal{A},s_\mathcal{B}^{\,\text{CT}},s_\mathcal{B}^{\,\mbox{\footnotesize $\mu$}}),
\end{align*}
where ${s_\mathcal{B}^{\,\text{CT}}=\text{MLP}_\mathcal{B}^{\,\text{CT}}(s_\mathcal{B})}$ and ${s_\mathcal{B}^{\,\mbox{\footnotesize $\mu$}}=\text{MLP}_\mathcal{B}^{\,\mbox{\footnotesize $\mu$}}(s_\mathcal{B})}$. MLP denotes a multi-layer perceptron composed of several linear layers with a non-linear activation after each layer. Additionally, we set a learnable parameter ${\alpha}$ such that the networks can determine how much of the style to apply considering that the amount of the style information the networks require may vary, i.e., ${c_{\mathcal{A}\rightarrow{\mathcal{B}}}=\alpha(\text{GDWCT}(c_\mathcal{A},s_\mathcal{B}^{\,\text{CT}},s_\mathcal{B}^{\,\mbox{\footnotesize $\mu$}}))+(1-\alpha)c_\mathcal{A}}$.

\begin{figure}[t]
\vspace*{-0.9cm}
  \includegraphics[width=\linewidth]{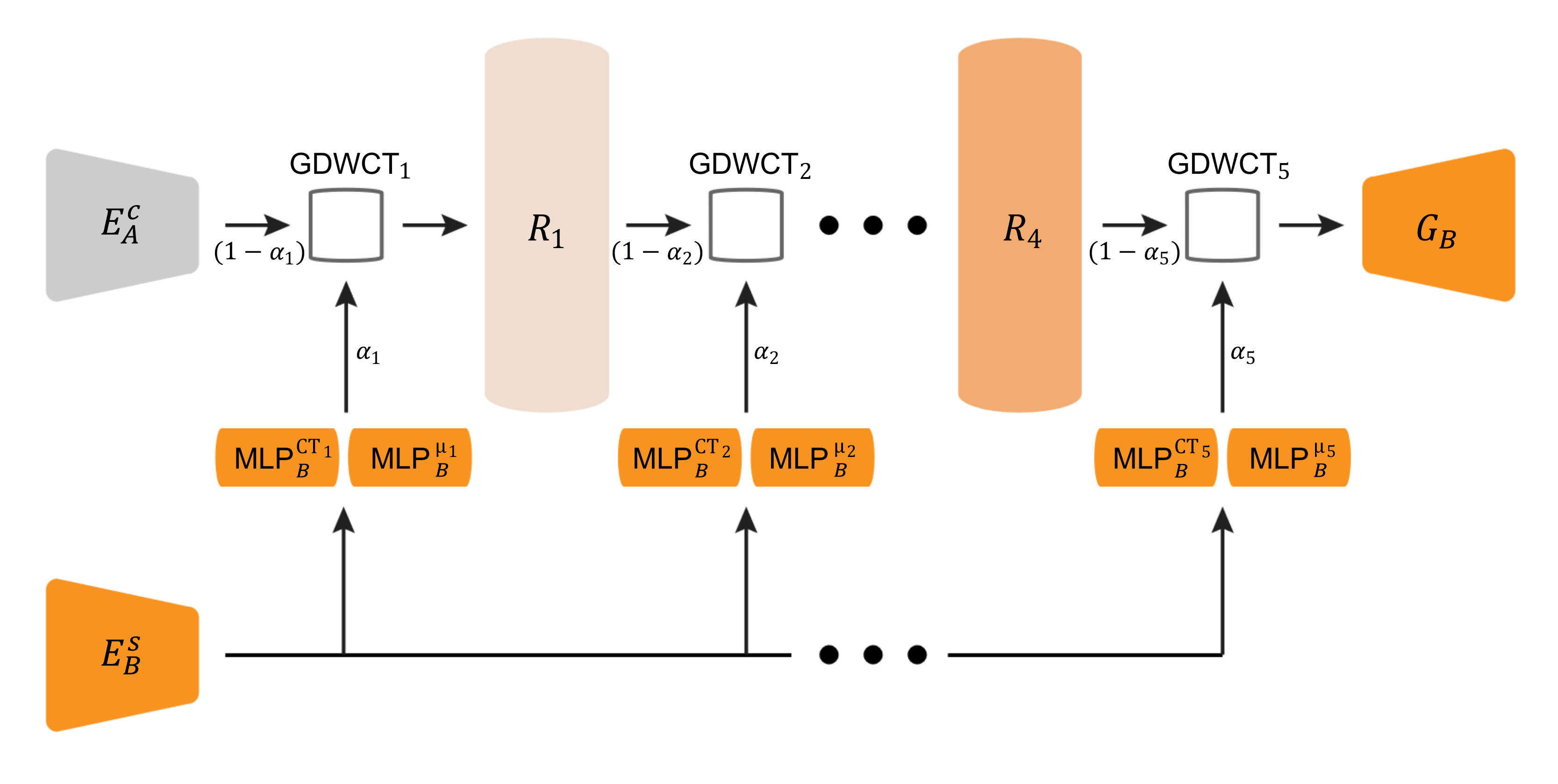}
  \vspace*{-0.8cm}
  \caption{Image translation via the proposed GDWCT. We apply the style via multiple hops to apply the style from the low-level feature to the high-level feature.}
\label{gdwct_fig}
\vspace*{-0.3cm}
\end{figure}

The different layers of a model focus on different information (e.g., the low-level feature captures a local fine pattern, whereas the high-level one captures a complicated pattern across a wide area). We thus add our GDWCT module in each residual block ${R_i}$ of the generator ${G_\mathcal{B}}$ as shown in Fig.~\ref{gdwct_fig}. By injecting the style information across multiple hops via a sequence of GDWCT modules, our model can simultaneously reflect both the fine- and coarse-level style information.




\subsection{Loss Functions}\label{loss}
Following MUNIT~\cite{Huang_2018_ECCV} and DRIT~\cite{Lee_2018_ECCV}, we adopt both the latent-level and the pixel-level reconstruction losses. First, we use the style consistency loss between two style features ${(s_{\mathcal{A}\rightarrow{\mathcal{B}}}, s_\mathcal{B})}$, so that it encourages the model to reflect the style of the reference image ${s_\mathcal{B}}$ to the translated image ${x_{\mathcal{A}\rightarrow{\mathcal{B}}}}$, i.e., 
\begin{align*}
    \mathcal{L}_{s}^{\mathcal{A}\rightarrow \mathcal{B}}=\mathop{{}\mathbb{E}_{x_{\mathcal{A}\rightarrow \mathcal{B}},x_\mathcal{B}}}~[\lVert E_{\mathcal{B}}^s(x_{\mathcal{A}\rightarrow \mathcal{B}})-E_\mathcal{B}^s(x_\mathcal{B}) \rVert_1]
\end{align*}
Second, we utilize the content consistency loss between two content features $({c_\mathcal{A}, c_{\mathcal{A}\rightarrow{\mathcal{B}}})}$ to enforce the model to maintain the content feature of the input image ${c_\mathcal{A}}$ after being translated ${c_{\mathcal{A}\rightarrow{\mathcal{B}}}}$, i.e. ,
\begin{align*}
    \mathcal{L}_{c}^{\mathcal{A}\rightarrow \mathcal{B}}=\mathop{{}\mathbb{E}_{x_{\mathcal{A}\rightarrow \mathcal{B}},x_\mathcal{A}}}[\lVert E_\mathcal{B}^c(x_{\mathcal{A}\rightarrow \mathcal{B}})-E_\mathcal{A}^c(x_\mathcal{A}) \rVert_1]
\end{align*}
Third, in order to guarantee the performance of our model through pixel-level supervision, we adopt the cycle consistency loss and the identity loss~\cite{Zhu_2017} to obtain a high-quality image, i.e.,
\begin{align*}
    \mathcal{L}_{cyc}^{\mathcal{A}\rightarrow \mathcal{B}\rightarrow \mathcal{A}}&=
    \mathop{{}\mathbb{E}_{x_\mathcal{A}}}\left[\lVert x_{\mathcal{A}\rightarrow \mathcal{B}\rightarrow \mathcal{A}}-x_\mathcal{A}\rVert_1\right] \\
    \mathcal{L}_{i}^{\mathcal{A}\rightarrow \mathcal{A}}&=
    \mathop{{}\mathbb{E}_{x_\mathcal{A}}}\left[\lVert x_{\mathcal{A}\rightarrow \mathcal{A}}-x_\mathcal{A}\rVert_1\right].
\end{align*}
Lastly, we use an adversarial loss for minimizing the discrepancy between the distribution of the real image and that of the generated image. In particular, we employ LSGAN~\cite{mao2017least} as the adversarial method, i.e.,
\begin{align*}
    \mathcal{L}_{D_{adv}}^{\mathcal{B}}&=\tfrac{1}{2}\mathop{{}\mathbb{E}_{x_\mathcal{B}}}[(D(x_\mathcal{B})-1)^2]+\tfrac{1}{2}\mathop{{}\mathbb{E}_{x_{\mathcal{A}\rightarrow{\mathcal{B}}}}}[(D(x_{\mathcal{A}\rightarrow{\mathcal{B}}}))^2]\\
    \mathcal{L}_{G_{adv}}^{\mathcal{B}}&=\tfrac{1}{2}\mathop{{}\mathbb{E}_{x_{\mathcal{A}\rightarrow{\mathcal{B}}}}}[(D(x_{\mathcal{A}\rightarrow{\mathcal{B}}})-1)^2]
\end{align*}

To consider the opposite translation, similar to DRIT~\cite{Lee_2018_ECCV}, our model is trained in both directions, ${(\mathcal{A}\rightarrow{\mathcal{B}}\rightarrow{\mathcal{A}})}$ and ${(\mathcal{B}\rightarrow{\mathcal{A}}\rightarrow{\mathcal{B}})}$, at the same time. Finally, our full loss function is represented as
\begin{align*}
    \mathcal{L}_D&=\mathcal{L}_{D_{adv}}^\mathcal{A}+\mathcal{L}_{D_{adv}}^\mathcal{B}\\
    \mathcal{L}_G&=\mathcal{L}_{G_{adv}}^\mathcal{A}+\mathcal{L}_{G_{adv}}^\mathcal{B}+\lambda_{latent}(\mathcal{L}_s+\mathcal{L}_c)+\\  
    &\quad\;\lambda_{pixel}(\mathcal{L}_{cyc}+\mathcal{L}_i^{\mathcal{A}\rightarrow{\mathcal{A}}}+\mathcal{L}_i^{\mathcal{B}\rightarrow{\mathcal{B}}})
\end{align*}
where ${\mathcal{L}}$ without a domain notation indicates both directions between two domains, and we empirically set ${\lambda_{latent}=1}$, ${\lambda_{pixel}=10}$.

\begin{figure*}
\vspace*{-0.9cm}
  \includegraphics[width=\linewidth]{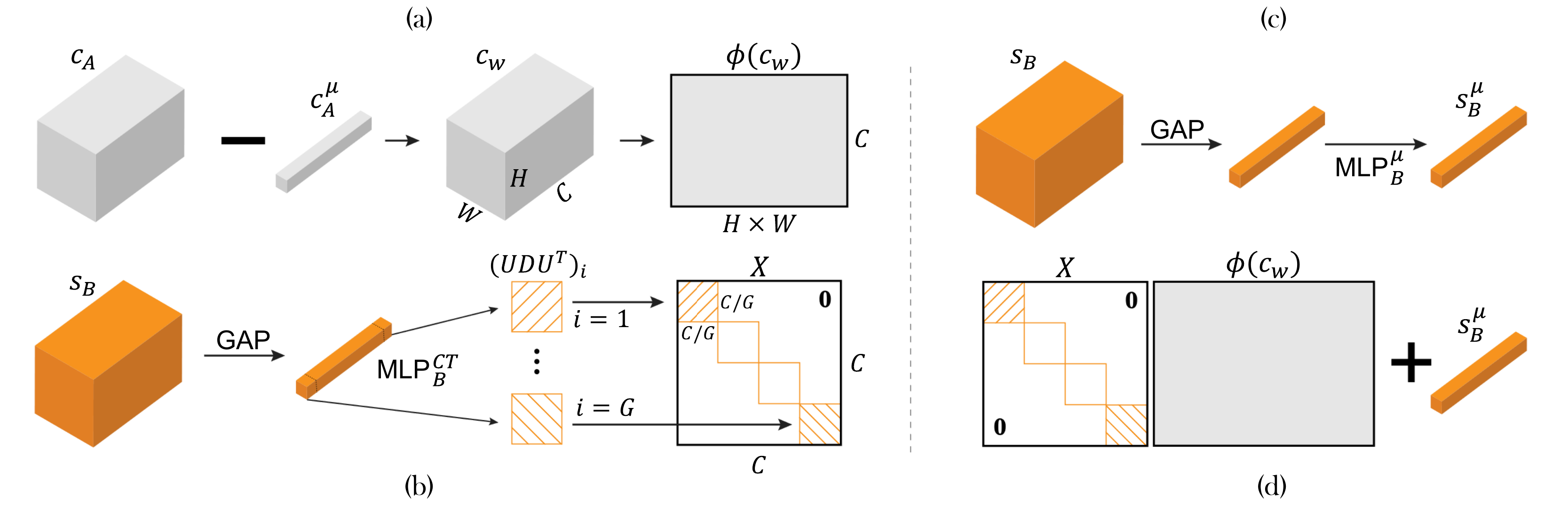}
  \vspace*{-0.7cm}
  \caption{Details on the proposed GDWCT module. (a) The process for obtaining the whitened feature. Because the regularization (Eq.~\eqref{r_w}) encourages the zero-mean content feature ${c-c^{{\,\mbox{\footnotesize $\mu$}}}}$ to be the whitened feature ${c_w}$, we just subtract the mean of the content feature ${c_\mathcal{A}^{\,\mbox{\footnotesize $\mu$}}}$ from ${c_\mathcal{A}}$. (b) The procedure of approximating the coloring transformation matrix (Section~\ref{gdwct}). (c) We obtain the mean of the style feature ${s_\mathcal{B}^{\,\mbox{\footnotesize $\mu$}}}$ by forwarding it to the MLP layer ${\text{MLP}_{\mathcal{B}}^{\,\mbox{\footnotesize $\mu$}}}$. (d) Our module first multiply the whitened feature ${c_w}$ with the group-wise coloring transformation matrix $X$. We then add it with the mean of the style ${s_\mathcal{B}^{\,\mbox{\footnotesize $\mu$}}}$.}
\label{gdwct_detail_fig}
\vspace*{-0.1cm}
\end{figure*}

\subsection{Group-wise Deep Whitening-and-Coloring Transformation} \label{gdwct}
For concise expression, we omit the domain notation unless needed, such as ${c}=\{{c_\mathcal{A},c_\mathcal{B}}\}$, ${s}=\{{s_\mathcal{A},s_\mathcal{B}}\}$, etc.

\paragraph{Whitening transformation (WT).} WT is a linear transformation that makes the covariance matrix of a given input into an identity matrix. Specifically, 
we first subtract the content feature ${c\in\mathcal{R}^{C\times BHW}}$ by its mean ${c^{\,\mbox{\footnotesize $\mu$}}}$, where ${(C,B,H,W)}$ represent the number of channels, batch size, height, and width, respectively. We then compute the outer product of the zero-meaned ${c}$ along the ${BHW}$ dimension. Lastly, we obtain the covariance matrix ${\Sigma_c\in\mathcal{R}^{C\times C}}$ and factorize it via eigendecomposition, i.e.,
\begin{align*}
    \Sigma_c=\tfrac{1}{{BHW}-1}\Sigma_{i=1}^{BHW}(c_i-c^{\,\mbox{\footnotesize $\mu$}})(c_i-c^{{\,\mbox{\footnotesize $\mu$}}})^T=Q_c\Lambda_c Q_c^T,
\end{align*}
where $Q_c\in\mathcal{R}^{C\times C}$ is the orthogonal matrix containing the eigenvectors, and ${\Lambda_c\in\mathcal{R}^{C\times C}}$ indicates the diagonal matrix of which each diagonal element is the eigenvalue corresponding to each column vector of ${Q_c}$. 
The whitening transformation is defined as
\begin{equation}\label{wt2}
    c_{w}=Q_c\Lambda_c^{-\frac{1}{2}}Q_c^T(c-c^{{\,\mbox{\footnotesize $\mu$}}}),
\end{equation}
where ${c_w}$ denotes the whitened feature. However, as pointed out in Section~\ref{intro}, eigendecomposition is not only computationally intensive but also difficult to backpropagate the gradient signal. To alleviate the problem, we propose the deep whitening transformation (DWT) approach such that the content encoder ${E_c}$ can naturally encode the whitened feature ${c_{w}}$, i.e., ${c_{w}=c-c^{{\,\mbox{\footnotesize $\mu$}}}}$,
where ${E_c(x_c)=c}$.
To this end, we propose the novel regularization term that makes the covariance matrix of the content feature $\Sigma_c$ as close as possible to the identity matrix, i.e.,
\begin{equation}\label{r_w}
    \mathcal{R}_{w}=
    \mathop{{}\mathbb{E}}[\lVert \Sigma_c-I \rVert_{1,1}].
\end{equation}
Thus, the whitening transformation in Eq.~\eqref{wt2} is reduced to ${c_{w}=c-c^{{\,\mbox{\footnotesize $\mu$}}}}$ in DWT.


However, several limitations exist in DWT. First of all, estimating the full covariance matrix using a small batch of given data is inaccurate~\cite{huang2018decorrelated}. Second, performing DWT with respect to the entire channels may excessively throw away the content feature, compared to channel-wise standardization. We therefore improve DWT by grouping channels and applying DWT to the individual group. 

Concretely, the channel dimension of ${c}$ is re-arranged at a group level, i.e., ${c\in {\mathcal{R}^{G\times (C/G)\times BHW}}}$, where ${G}$ is the number of groups. After obtaining the covariance matrix ${\Sigma_c}$ in $\mathcal{R}^{G\times (C/G)\times (C/G)}$, we apply Eq.~\eqref{r_w} along its group dimension. Note that group-wise DWT (\textbf{GDWT}) is the same with DWT during the forward phase, as shown in Fig.~\ref{gdwct_detail_fig}(a), because the re-arranging procedure is required for the regularization~\eqref{r_w}. 

\vspace*{-0.3cm}
\paragraph{Coloring transformation (CT).} CT matches the covariance matrix of the whitened feature with that of the style feature ${\Sigma_s}$, where ${\Sigma_s}$ is the covariance matrix of the style feature. ${\Sigma_s}$ is then decomposed into ${Q_s\Lambda_sQ_s^T}$, used for the subsequent coloring transformation. This process is written as  
\begin{equation}\label{ct}
    \begin{matrix}
        c_{cw}=Q_s\Lambda_s^{\frac{1}{2}}Q_s^T c_{w},
    \end{matrix}
\end{equation}
where ${c_{cw}}$ denotes the colored feature. 

Similar to WT, however, CT has the problems of expensive time complexity and non-trivial backpropagation. Thus, We also replace CT with a simple but effective method that we call a deep coloring transformation (DCT). Specifically, we first obtain the matrix ${s^{\text{CT}}}$ through ${\text{MLP}^{\,\text{CT}}(s)}$, where ${s=E_s(x)}$. We then decompose ${s^{\text{CT}}}$ into two matrices by computing its column-wise $L_2$ norm, i.e., ${s^{\text{CT}}=UD}$, where the ${i}$-th column vector ${u_i}$ of $U\in\mathcal{R}^{C\times C}$ is the unit vector, and $D\in\mathcal{R}^{C\times C}$ is the diagonal matrix whose diagonal entries correspond to the $L_2$ norm of each column vector of ${s^{\text{CT}}}$. We assume that those matrices ${UD}$ is equal to two matrices in Eq.~\eqref{ct}, i.e., ${UD=}$ ${Q_s\Lambda_s^{\frac{1}{2}}}$.

In order to properly work as ${Q_s}$ and ${\Lambda_s^{\frac{1}{2}}}$, ${U}$ needs to be an orthogonal matrix, and every diagonal entry in the matrix ${D}$ should be positive. To assure the conditions, we add the regularization for ${U}$ to encourage the column vectors of ${U}$ to be orthogonal, i.e., 
\begin{equation}\label{r_c}
    \mathcal{R}_{c}=
    \mathop{{}\mathbb{E}_{s}}~[\lVert U^TU-I \rVert_{1,1}].
\end{equation}
The diagonal matrix ${D}$ has its diagonal elements as the column-wise ${L_2}$ norm of ${s^\text{CT}}$, such that its diagonal entries are already positive. Thus, it does not necessitate additional regularization. Meanwhile, ${U}$ becomes the orthogonal matrix if ${U}$ accomplishes the orthogonality, because each column vector ${u_i}$ of ${U}$ has a unit $L_2$ norm. That is, with the regularization Eq.~\eqref{r_c}, ${UD}$ satisfies the entire conditions to be ${Q_s\Lambda_s^{\frac{1}{2}}}$.
Finally, combining ${U}$ and ${D}$, we simplify CT as 
\begin{equation}\label{dct}
    c_{cw}=UDU^Tc_w.
\end{equation}

However, approximating the entire matrix ${s^{\text{CT}}}$ has an expensive computational cost (the number of parameters to estimate is ${C^2}$). Hence, we extend DCT to the group-wise DCT (\textbf{GDCT}) and reduce the number of parameters from ${{C^2}}$ to ${C^2/G}$, as the detailed steps are illustrated in Fig.~\ref{gdwct_detail_fig}(b). We first obtain the ${i}$-th matrix $\{{UDU^T}\}_i\in\mathcal{R}^{(C/G)\times (C/G)}$ for GDCT for ${i=\{1,...,G\}}$. We then form a block diagonal matrix ${X\in\mathcal{R}^{C\times C}}$ by arranging the matrices ${\{UDU^T\}_{1,...,G}}$. Next, as shown in Fig.~\ref{gdwct_detail_fig}(d), we compute the matrix multiplication with ${X}$ and the whitened feature ${c_w}$, thus Eq.~\eqref{dct} being reduced to
\begin{align*}
    c_{cw} = X\phi(c_w),
\end{align*}
where ${\phi}$ denotes a reshaping operation ${\phi:\mathcal{R}^{C\times H\times W}\rightarrow}$ ${{\mathcal{R}^{C\times HW}}}$. Finally, we add the new mean vector ${s^{\,\mbox{\footnotesize $\mu$}}}$ to the ${c_{cw}}$, where ${s^{\,\mbox{\footnotesize $\mu$}}=\text{MLP}^{\,\mbox{\footnotesize $\mu$}}(s)}$, as shown in Fig.~\ref{gdwct_detail_fig}(c). We empirically set ${\lambda_{w}=0.001,\lambda_{c}=10}$, and ${G=4,8,16}$.



\begin{figure*}[t]
\vspace*{-0.9cm}
  \includegraphics[width=\linewidth]{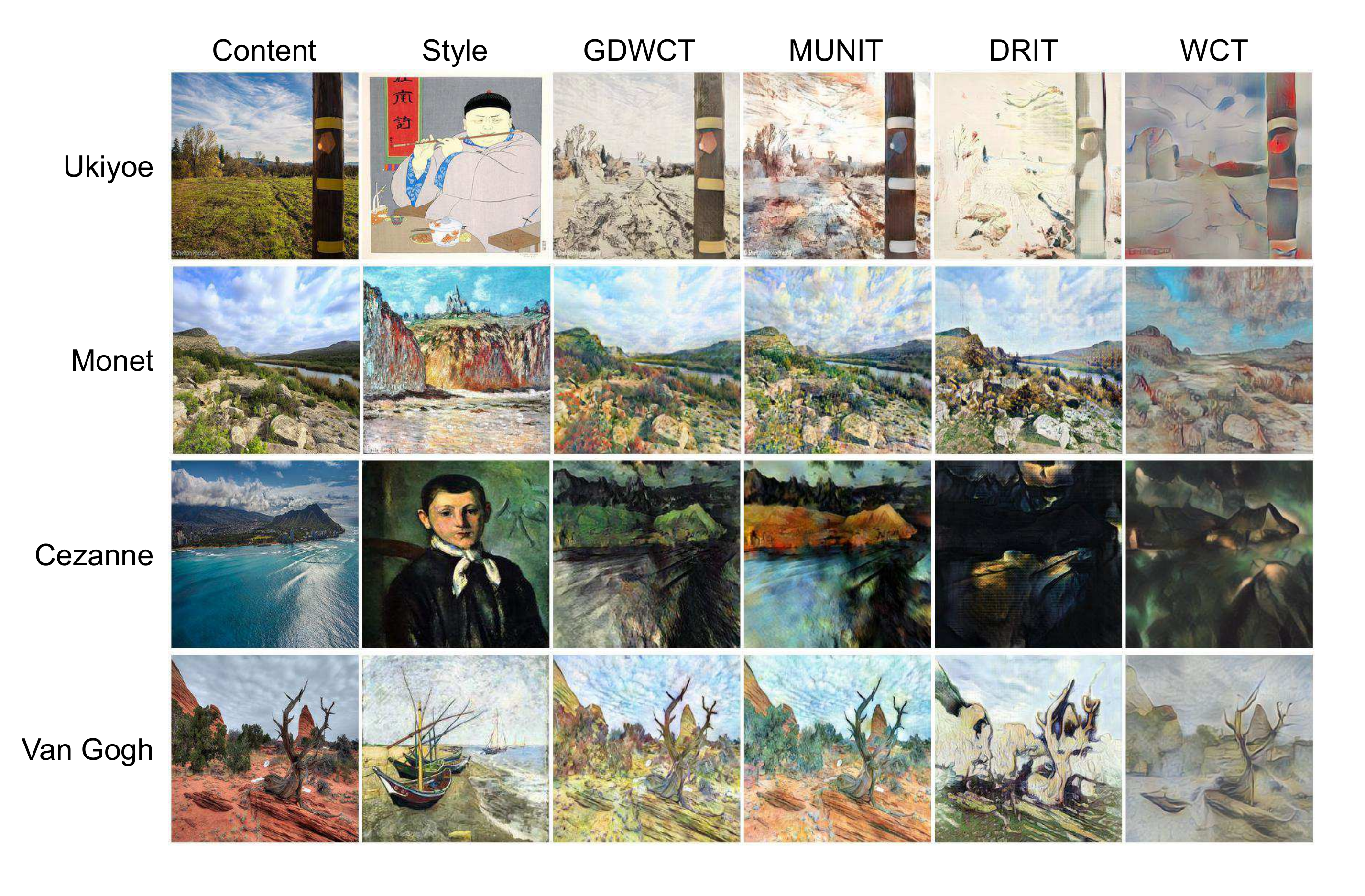}
  \vspace*{-1.2cm}
  \caption{Qualitative comparisons based on Artworks dataset~\cite{Zhu_2017}.}
\label{art}
  \vspace*{-0.0cm}
\end{figure*}

\section{Experiments}\label{experiment}
This section describes the baseline models and the datasets. Implementation details as well as additional comparisons and results are included in the appendix.

\subsection{Experimental Setup}\label{baseline}
\paragraph{Datasets.} 
We evaluate GDWCT with various datasets including CelebA~\cite{liu2015faceattributes}, Artworks~\cite{Zhu_2017} (Ukiyoe, Monet, Cezanne, and Van Gogh), cat2dog~\cite{Lee_2018_ECCV}, Pen ink and Watercolor classes of the Behance Artistic Media (BAM)~\cite{Wilber_2017_ICCV}, and Yosemite~\cite{Zhu_2017} (summer and winter scenes) datasets.

\paragraph{Baseline methods.} 
We exploit MUNIT~\cite{Huang_2018_ECCV}, DRIT~\cite{Lee_2018_ECCV}, and WCT~\cite{li2017universal} as our baselines because those methods are the state-of-the-art in image translation and style transfer, respectively. MUNIT and DRIT utilize different methods when applying the style into the content from GDWCT. MUNIT leverages AdaIN~\cite{huang2017arbitrary} while DRIT is based on concatenation of the content and the style features. Meanwhile, WCT applies the whitening-and-coloring transformation to the features extracted from the pretrained encoder, in order to transfer the style into the content image. 

\subsection{Quantitative Analysis}\label{quan}
We compare the performance of our model with the baselines with user study and classification accuracy. 

\begin{table}[!b]
\begin{center}
\small
\vspace*{-0.2cm}
\begin{tabular}{ccccc}
\toprule
&MUNIT & DRIT & WCT & GDWCT\\
\midrule
Male $\rightarrow$ Female & 4.41 & 42.25 & 10.12 & \textbf{44.52}\\
Female $\rightarrow$ Male & 7.78 & \textbf{48.89} & 4.44 & 38.89\\
Bang $\rightarrow$ Non-Bang & 3.35 & 42.20 & 3.37 & \textbf{51.10}\\
Non-Bang $\rightarrow$ Bang & 6.67 & 18.89 & 4.45 & \textbf{71.15}\\
Smile$\rightarrow$ Non-Smile & 5.56 & 30.35 & 1.35 & \textbf{64.44}\\
Non-Smile$\rightarrow$ Smile & 2.30 & 22.25 & 2.25 & \textbf{73.33}\\ 
\bottomrule
\end{tabular}
\end{center}
\vspace*{-0.5cm}
\caption{Comparisons on the user preference. Numbers
indicate the percentage of preference on each class. }
\label{userstudy_tab}
\vspace*{-0.3cm}
\end{table}

\paragraph{User study.}
We first conduct a user study using CelebA dataset~\cite{liu2015faceattributes}. The participants of the user study was instructed to measure user preferences on outputs produced by GDWCT and the baseline models with a focus on the quality of an output and the rendering of the style given in an exemplar. Each user evaluated 60 sets of image comparisons, choosing one among four candidates within 30 seconds per comparison. We informed the participants of the original and the target domains for every run, e.g., Male to Female, so that they can understand exactly which style in an exemplar is of interest. Table~\ref{userstudy_tab} summarizes the result. It is found that the users prefer our model to other baseline models on five out of six class pairs. In the translation of (${\text{Female}\rightarrow{\text{Male}}}$), because DRIT consistently generates a facial hair in all translation, it may obtain the higher score than ours. The superior measures demonstrate that our model produces visual compelling images. Furthermore, the result indicates that our model reflect the style from the exemplar better than other baselines, which justifies that matching entire statistics including a covariance would render style more effectively. 

\paragraph{Classification accuracy.} A well-trained image translation model would generate outputs that are classified as an image from the target domain. For instance, when we translate a female into male, we measure the classification accuracy in the gender domain. A high accuracy indicates that the model learns deterministic patterns to be represented in the target domain. The classification results are reported in Table~\ref{ca_tab}. For the classification, we adopted the pretrained Inception-v3~\cite{szegedy2016rethinking} and fine-tuned on CelebA dataset. Our model records competitive average on the accuracy rate, marginally below DRIT on Gender class, and above on Bangs and Smile.

\begin{table}[h]
\begin{center}
\small
\begin{tabular}{ccccc}
\toprule
& MUNIT & DRIT & WCT & GDWCT \\
\midrule
Gender & 30.10 & \textbf{95.55} & 28.80 & 92.65 \\
Bangs & 35.43 & 66.88 & 24.85 & \textbf{76.05} \\
Smile & 45.60 & 78.15 & 32.08 & \textbf{92.85} \\
\midrule
Avg. & 37.04 & 80.19 & 28.58 & \textbf{87.18}  \\ 
\bottomrule
\end{tabular}
\end{center}
 \vspace*{-0.3cm}
\caption{Comparison of the classification accuracy (\%).}
\label{ca_tab}
\vspace*{-0.3cm}
\end{table}

\paragraph{Inference time.} 
The superiority of GDWCT also lies in the speed at which outputs are computed in the inference stage. Table~\ref{runtime_tab} shows that our model is as fast as the existing image translation methods, and has the capacity of rendering rich style information as of WCT. The numbers represent the time taken to generate one image. 

\begin{table}[h]
\begin{center}
\small
\begin{tabular}{ccccc}
\toprule
& MUNIT & DRIT & WCT & GDWCT \\
\midrule
\begin{tabular}{@{}c@{}}Runtime (sec)\end{tabular} & 0.0419 & 0.0181 & 0.8324 & 0.0302\\
\bottomrule
\end{tabular}
\end{center}
\vspace*{-0.5cm}
\caption{Comparison of the inference time. Tested with the image size 256$\times$256 on a NVIDIA Titan XP GPU, and averaged over 1,000 trials.}
\label{runtime_tab}
\vspace*{-0.3cm}
\end{table}

\subsection{Qualitative Results}\label{qual}
In this section, we analyze the effects of diverse hyperparameters and devices on the final image outputs.

\begin{figure}[t]
  \vspace*{-0.6cm}
  \includegraphics[width=\linewidth]{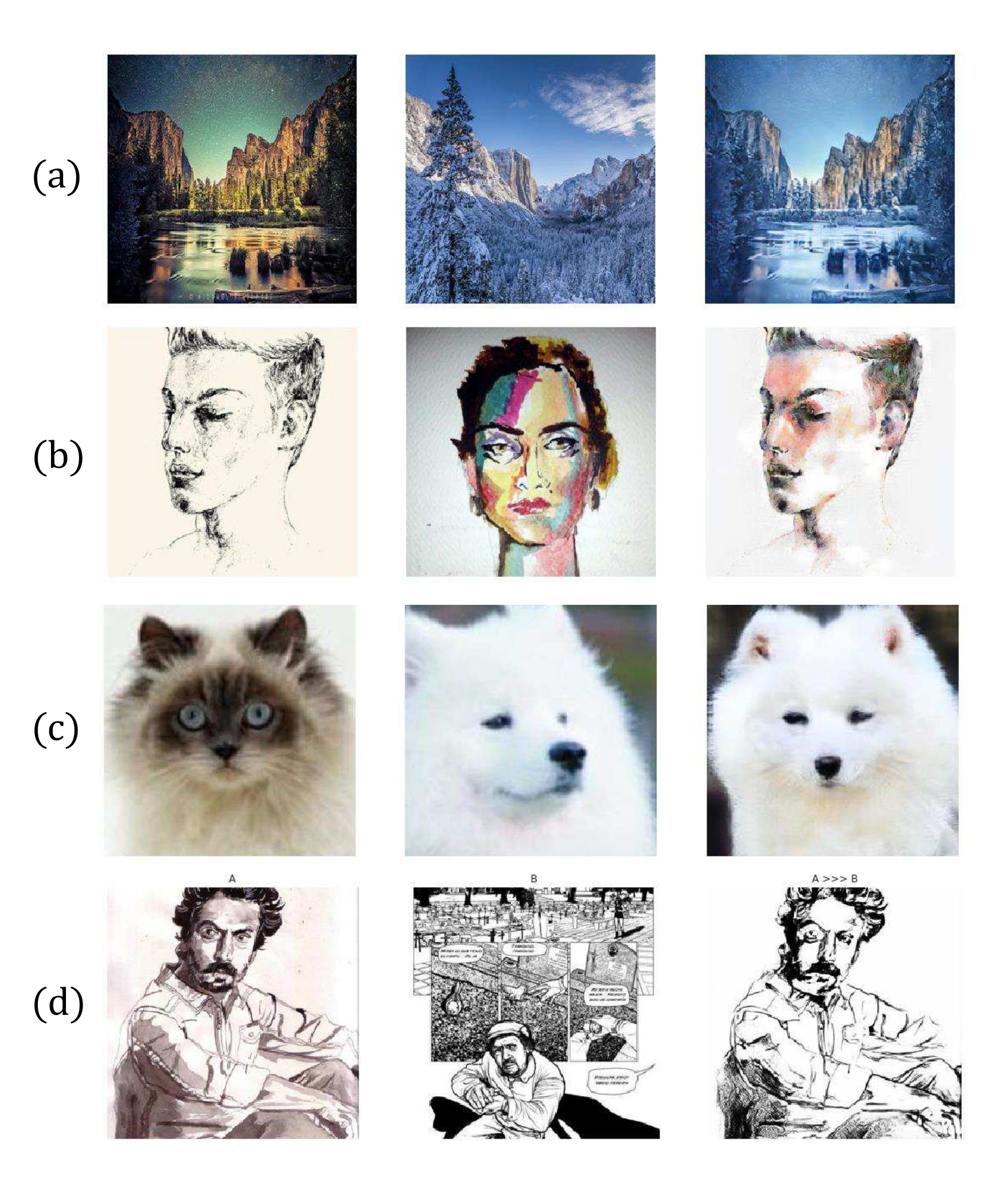}
  \vspace*{-0.7cm}
  \caption{Results on various dataset; (a) Yosemite (b) BAM (Pen Ink $\rightarrow$ Water Color) (c) Cat2dog (d) BAM (Water Color $\rightarrow$ Pen Ink)}
\label{diverse_results}
\vspace*{-0.3cm}
\end{figure}

\vspace*{-0.3cm}
\paragraph{Stylization comparisons.}
We conduct qualitative analyses by a comparison with the baseline models on Fig.~\ref{art}. Each row represents different classes, and the leftmost and the second columns are content and the exemplar style, respectively. Across diverse classes, we observe consistent patterns for each baseline model. First, MUNIT tends to keep the object boundary, leaving not much room for style to get in. DRIT shows results of high contrast, and actively transfer the color. WCT is more artistic in the way it digests the given style, however at times losing the original content to a large extent. Our results transfer object colors as well as the overall mood in the style, while not overly blurring details. We provide additional results of our model in Fig.~\ref{diverse_results}. We believe our work gives another dimension of an opportunity to translate the image at one's discretion.

\begin{figure}[!b]
\vspace*{-0.2cm}
  \includegraphics[width=\linewidth]{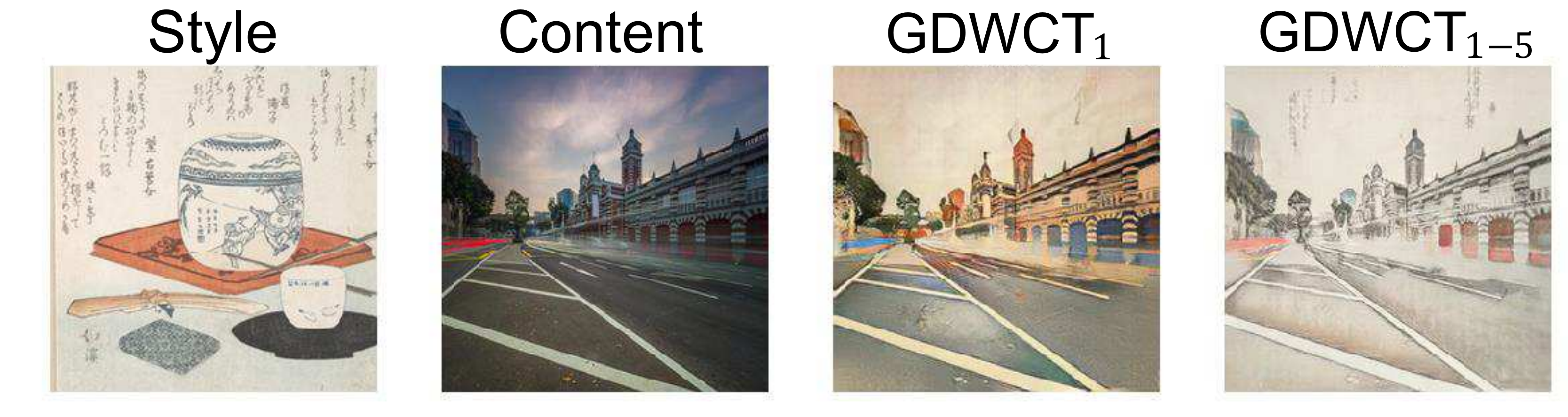}
  \vspace*{-0.4cm}
  \caption{Comparison between single- and multi-hops.}
\label{single_multi}
\vspace*{-0.1cm}
\end{figure}

\vspace*{-0.3cm}
\paragraph{Number of hops on style.} As we previously discussed in Fig.~\ref{gdwct_fig}, the proposed GDWTC could be applied in multi-hops. We demonstrate the effects of the different number of hops on the style. To this end, we use Artworks dataset (Ukiyoe)~\cite{Zhu_2017}. We train two identical models different only in the number of hops, a single hop (GDWTC$_1$) or multi-hops ($\text{GDWTC}_{1-5}$). In Fig.~\ref{single_multi}, the rightmost image ($\text{GDWTC}_{1-5}$) has the style that agrees with the detailed style given in the leftmost image. The third image (GDWTC$_1$) follows the overall color pattern of the exemplar, but with details less transferred. For example, the writing in the background has not been transferred to the result of GDWCT$_1$, but is clearly rendered on $\text{GDWTC}_{1-5}$. The difference comes from a capacity of the multiple hops on a stylization, which covers both fine and coarse style~\cite{li2017universal}.

\begin{figure}[t]
  \includegraphics[width=\linewidth]{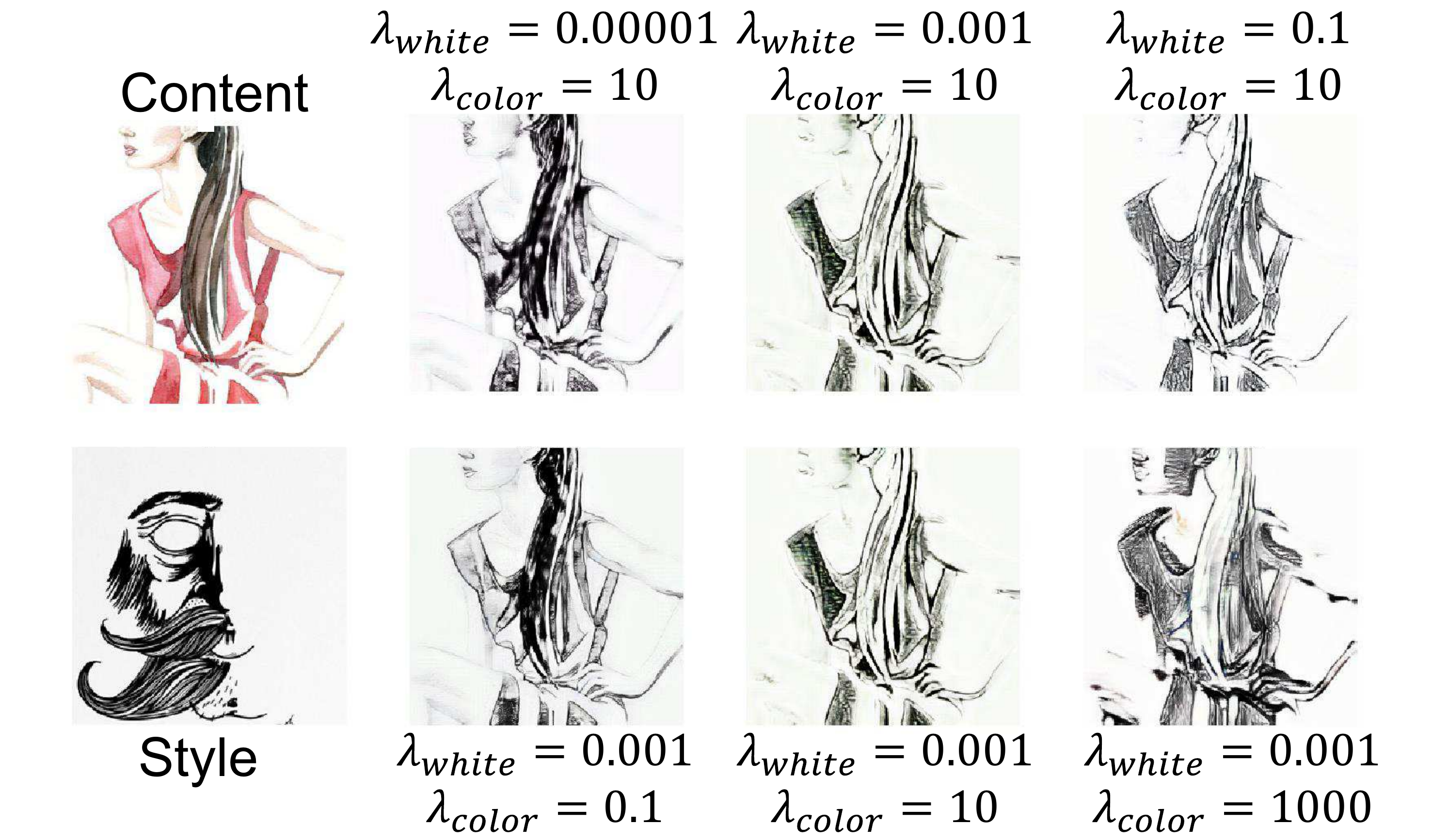}
 \vspace*{-0.5cm}
  \caption{Visualization of the regularization influences.}
\label{ablation}
\vspace*{-0.3cm}
\end{figure}

\begin{figure*}[!b]
  \includegraphics[width=\linewidth]{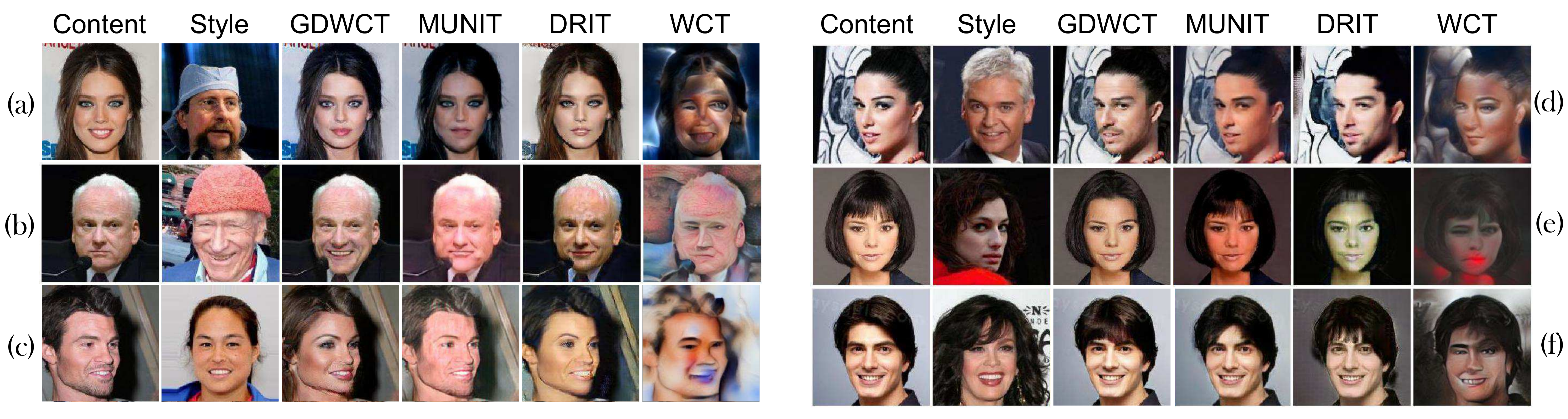}
  \vspace*{-0.7cm}
  \caption{Comparison with the baseline models on CelebA dataset; (a) ${\text{Smile}\rightarrow{\text{Non-Smile}}}$ (b) ${\text{Non-Smile}\rightarrow{\text{Smile}}}$ \\(c) ${\text{Male}\rightarrow{\text{Female}}}$ (d) ${\text{Female}\rightarrow{\text{Male}}}$ (e) ${\text{Bang}\rightarrow{\text{Non-Bang}}}$ (f) ${\text{Non-Bang}\rightarrow{\text{Bang}}}$}
\label{fig:comparison with the baseline models}
\vspace*{-0.3cm}
\end{figure*}

\vspace*{-0.3cm}
\paragraph{Effects of regularization.} We verify the influences of the regularizations ${\mathcal{R}_{w}}$ and ${\mathcal{R}_{c}}$ on the final image output. Intuitively, a higher ${\lambda_{w}}$ will strengthen the whitening transformation, erasing the style more, because it encourages the covariance matrix of the content feature to be closer to the identity matrix. Likewise, a high value of ${\lambda_{c}}$ would result in a diverse level of style, since the intensity of the style applied during coloring increases as the eigenvectors of the style feature gets closer to orthogonal. 

We use two classes, Watercolor and Pen Ink, of BAM~\cite{Wilber_2017_ICCV} dataset. The images in Fig.~\ref{ablation} illustrates the results of ${(\text{Water color}\rightarrow{\text{Pen ink}})}$. Given the leftmost content and style as input, the top row shows the effects of gradually increasing value of ${\lambda_{w}}$. A large ${\lambda_{w}}$ leads the model to erase textures notably in the cloth and hair. It proves our presumption that the larger w is, the stronger the effects of the whitening is. Meanwhile, the second row shows the effects of different coloring coefficient ${\lambda_{c}}$. The cloth of the subjects shows a stark difference, gradually getting darker, applying the texture of the style more intensively.

\begin{figure}[t]
    \vspace*{-0.3cm}
  \includegraphics[width=\linewidth]{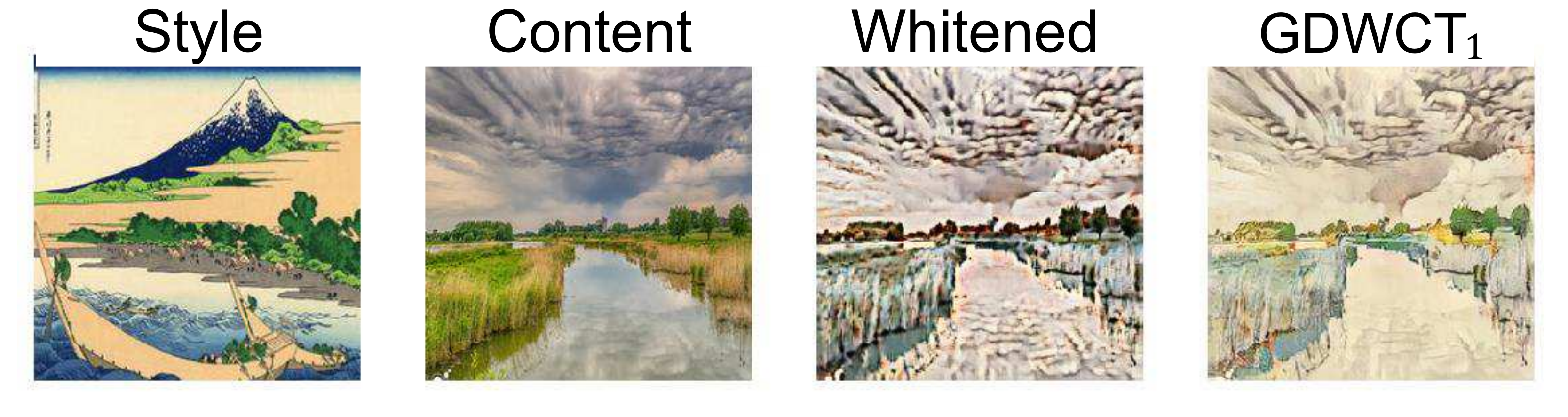}
  \vspace*{-0.6cm}
  \caption{Visualization of whitening transformation that makes the content feature lose the original information.}
\label{whitened}
\vspace*{-0.3cm}
\end{figure}

\vspace*{-0.3cm}
\paragraph{Visualization of whitening-and-coloring transformation.} We visualize the whitened feature to visually inspect the influence of the proposed group-wise deep whitening transformation on the content image. We also use a sample from Artworks dataset. For visualization, we forward the whitened feature into the networks without coloring transformation. The third image from the left shows the whitening effects. It is evident that in the image, detailed style regarding the color and texture are erased from the content image. Notably, the reeds around the river, and the clouds in the sky are found to be whitened in color, being ready to be stylized. On the other hand, the rightmost image stylizes given the whitened image via the group-wise deep coloring transformation. It reveals that the coloring transformation properly applies the exemplar style, which is in a simpler style with monotonous color than that of the content image. 

\vspace*{-0.5cm}
\paragraph{Comparison on face attribute translation.}\label{sub:additional comparison} We compare GDWCT with the baselines using CelebA dataset with the image size of 216${\times}$216. The results are shown in Fig.~\ref{fig:comparison with the baseline models}. Two columns from the left of each macro column denote a content image and a style image (exemplar), respectively, while the other columns indicates outputs of compared models. Each row of each macro column illustrates the different target attribute. Our model shows a superior performance in overall attribute translation, because our model drastically but suitably applies the style compared to the baselines. For example, In case of ${(\text{male}\rightarrow{\text{female}})}$ translation, our model generates an image with long hair and make-up, the major patterns of the woman. However, each generated image from MUNIT and DRIT wears only light make-up with incomplete long hair. Meanwhile, in both translation cases of Smile and Bangs, the outputs of MUNIT show less capacity than ours in transferring the style as shown in ${(\text{Smile}\rightarrow{\text{Non-Smile}})}$, ${(\text{Non-Bang}\rightarrow{\text{Bang}})}$, and ${(\text{Bang}\rightarrow{\text{Non-Bang}})}$, because MUNIT matches only mean and variance of the style to those of the content when conducting a translation. On the other hand, DRIT conducts unnatural translation (two rows from the bottom) comparing with ours. In case of ${(\text{Non-Smile}\rightarrow{\text{Smile}})}$, DRIT applies the style only into a mouth but ours converts both eyes and mouth. Meanwhile, as seen in overall cases of WCT, it cannot perform image translation because it does not learn to transfer the semantic style.

\section{Conclusion}
In this paper, we propose a novel framework, group-wise deep whitening-and-coloring transformation (GDWCT) for an improved stylization capability. Our experiments demonstrate that our work produces competitive outputs in image translation as well as style transfer domains, having a majority of real users agree that our model successfully reflects the given exemplar style. We believe this work bears the potential to enrich relevant academic fields with the novel framework and practical performances.   

\paragraph{Acknowledgement.} This work was partially supported by the National Research Foundation of Korea (NRF) grant funded by the Korean government (MSIP) (No. NRF2016R1C1B2015924). Jaegul Choo is the corresponding author.   

\clearpage

\section{Appendix}
In this section, we supplement our paper by reporting additional information. First of all, we describe the implementation details of our networks in subsection~\ref{sub:implementation}. We then provide a discussion on the effects of the number of groups with qualitative results in subsection~\ref{sub:effects of different number of groups}. Third, we qualitatively and quantitatively compare our model with the baseline models on CelebA dataset in subsection~\ref{sub:additional comparison with ELEGANT}.
Finally, we report extra results on CelebA dataset in subsection~\ref{sub:extra results}.

\subsection{Implementation}\label{sub:implementation}
\paragraph{Content encoder.} The content encoders $\{{E_\mathcal{A}^c,E_\mathcal{B}^c}\}$ are composed of a few strided convolutional (conv) layers and four residual blocks. The size of the output activation map is in ${\mathcal{R}^{256\times \frac{H}{4}\times \frac{W}{4}}}$. Note that we use the instance normalization~\cite{ulyanov2016instance} along with the entire layers in ${E_c}$ in order to flatten the content feature~\cite{Huang_2018_ECCV,huang2017arbitrary}. 


\paragraph{Style encoder.} The style encoders $\{{E_\mathcal{A}^s,E_\mathcal{B}^s}\}$ consist of several strided conv layers with the output size in ${\mathcal{R}^{256\times \frac{H}{16}\times \frac{W}{16}}}$. After the global average pooling, the style feature ${s}$ is forwarded into the ${\text{MLP}^{\,\text{CT}}}$ and $\text{MLP}^{\,\mbox{\footnotesize $\mu$}}$. We use the group normalization~\cite{Wu_2018_ECCV} in ${E_s}$ to match the structure of ${s}$ with ${\text{MLP}^{\,\text{CT}}}$ by grouping the highly correlated channels in advance.

\paragraph{Multi layer perceptron.}
Each of $\{{\text{MLP}_\mathcal{A}^{\,\text{CT}},\text{MLP}_\mathcal{B}^{\,\text{CT}}}\}$ and $\{{\text{MLP}_\mathcal{A}^{\,\mbox{\footnotesize $\mu$}},\text{MLP}_\mathcal{B}^{\,\mbox{\footnotesize $\mu$}}}\}$ is composed of several linear layers. The input dimension of ${\text{MLP}^{\,\text{CT}}}$ depends on the number of group. Specifically, the partial style feature in ${\mathcal{R}^{\frac{C}{G}}}$ is forwarded as the input feature and the output size is the square of the input dimension. On the other hand, both of the input and output dimension of $\text{MLP}^{\,\mbox{\footnotesize $\mu$}}$ is thesame with the number of channels, 256.

\paragraph{Generator.}
The generators $\{{G_\mathcal{A},G_\mathcal{B}\}}$ are made of four residual blocks and several sequence of upsampling layer with strided conv layer. Note that GDWCT is applied in the process of forwaring ${G}$.

\paragraph{Discriminator.}
The discriminators $\{{D_\mathcal{A},D_\mathcal{B}}\}$ are in the form of multi-scale discriminators~\cite{wang2018pix2pixHD}. The size of the output activations are in ${\mathcal{R}^{4\times 4},\mathcal{R}^{8\times 8}}$ and ${\mathcal{R}^{16\times 16}}$.
\paragraph{Training details.} We use the Adam optimizer~\cite{kinga2015method} with ${\beta_1=0.5,\beta_2=0.999}$ with a learning rate of 0.0001 for all generators and discriminators. Other settings are chosen differently based on the experimented dataset. In CelebA, we apply a batch size of eight with the image size of ${(216\times 216)}$. The original image size ${(178\times 218)}$ is resized to ${(216\times 264.5)}$, followed by the center-crop to be ${(216\times 216)}$. Corresponding models are trained for 500,000 iterations with a decaying rate of 0.5 applied from the 100,000th iteration in every 50,000 iterations. In all other datasets, We train the model with the batch size of two and the image size of ${(256\times 256)}$. Note that we first resize each image up to 286, then perform a random cropping. We set 200,000 iterations for the training and apply the decaying rate of 0.5 from 100,000th iterations in every 10,000 iterations. All the experiments are trained using a single NVIDIA TITAN Xp GPU for three days. The group size is empirically set among ${4,8,16}$.

\begin{figure}[t]
\begin{center}
  \includegraphics[width=\linewidth]{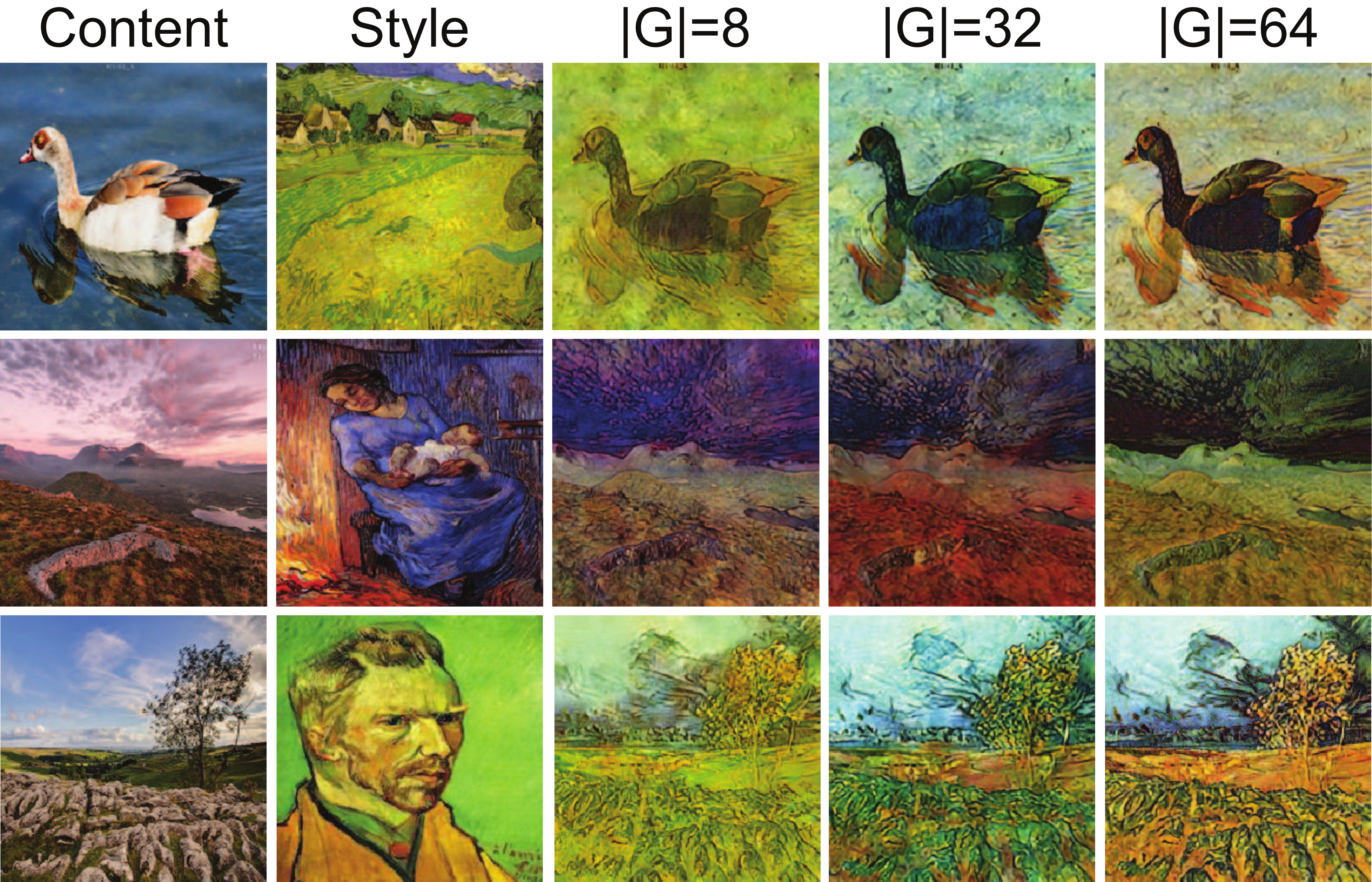}
\end{center}
  \caption{Effects of the number of groups. }
\label{fig:ablation on the number of groups.}
\end{figure}



\subsection{Effects of Different Number of Groups}\label{sub:effects of different number of groups}
We discuss and conduct additional experiments on the effects of different group sizes. 

First of all, the number of groups, ${|G|}$, is closely related to the number of model parameters to represent the style statistics of a given exemplar. Specifically, the number of model parameters is equivalent to ${C^{2}/|G|}$, where ${C}$ and ${|G|}$ represent the numbers of channels and groups, respectively, as discussed in Section 3.3. 
Thus, increasing ${|G|}$ has the effect of reducing the model size, i.e., the number of parameters. 

We also conduct a qualitative experiment to show the effects of ${|G|}$ on the final output. As shown in Fig.~\ref{fig:ablation on the number of groups.}, a small value of ${|G|}$ tends to focus on the low-level information of the style. For example, those results with ${|G|}=8$ in the third column mainly reflect the colors of the exemplar, while those results with ${|G|}=64$ in the rightmost column do not necessarily maintain the color tone. We additionally observe that the style becomes more distinguishable across different objects in a resulting image as ${|G|}$ increases, such that the color of the duck in the first row becomes more distinct from the background as ${|G|}$ gets larger. We believe it is ascribed to the fact that larger ${|G|}$ shows the better capability in capturing contrast between objects in the image. 

Although we attempted to rigorously figure out the effects of ${|\text{G}|}$ on our method, however, through several experiments, ${|\text{G}|}$ sometimes shows inconsistent patterns, so that the generalization of the effects of ${|\text{G}|}$ is vague and difficult. Thus, as a future work, it is required to explore in-depth the influences the number of groups gives rise to.

\subsection{Additional Comparison Results}\label{sub:additional comparison with ELEGANT}
\begin{figure}
\begin{center}
  \includegraphics[width=\linewidth]{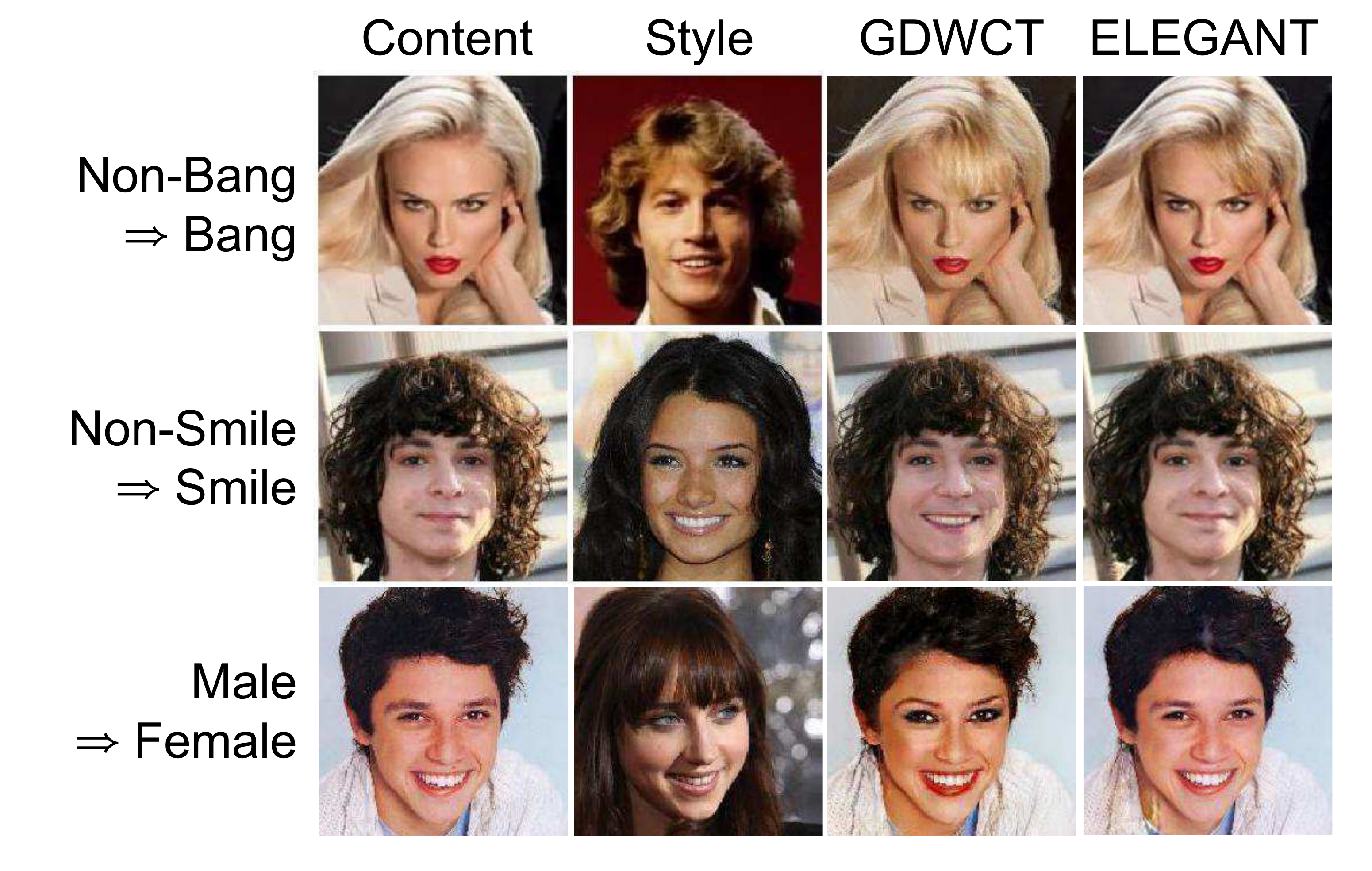}
\end{center}
  \caption{Qualitative comparison on attribute translation. Tested with image size of 216$\times$216.}
\label{fig:elegant_compare}
\end{figure}

In order to further validate the performances of our method, we additionally compare our method against ELEGANT~\cite{Xiao_2018_ECCV}, a recently proposed approach that focuses on facial attribute translation and exploits the adversarial loss. As shown in Fig.~\ref{fig:elegant_compare}, qualitative results show that our method performs better than ELEGANT in terms of intended attribute translation. For instance, in the first row, our method generates more luxuriant bangs than the baseline method when translating from `Non-Bang' to `Bang'. Better results are also found in the Smile attribute, which shows the results closer to the given style. The person in the last row is translated to a female of a high quality with regard to the eyes. ELEGANT encodes all target attributes in a disentangled manner in a single latent space and substitutes a particular part of the feature from one image to another. Since ELEGANT neither decomposes a given image into the content and the style nor matches the statistics of the style to that of the content, it shows worse performances in properly reflecting the style image than our proposed model.


Furthermore, we also show the outstanding performance of our method in a quantitative manner, as illustrated in Table~\ref{table: comparison on classification accuracy}. In all cases, our model achieves a higher classification accuracy by a large margin.

\begin{table}[t]
\begin{center}
\begin{tabular}{ccccc}
\toprule
& Gender & Bangs & Smile & Avg. \\
\midrule
ELEGANT & 77.15 & 61.73 & 70.88 & 69.92 \\
Ours & \textbf{92.65} & \textbf{76.05} & \textbf{92.85} & \textbf{87.18}  \\ 
\bottomrule
\end{tabular}
\end{center}
\caption{Classification accuracy in percentages.}
\label{table: comparison on classification accuracy}
\end{table}

\subsection{Extra Results}\label{sub:extra results} 
\begin{figure*}[t]
  \includegraphics[width=\linewidth]{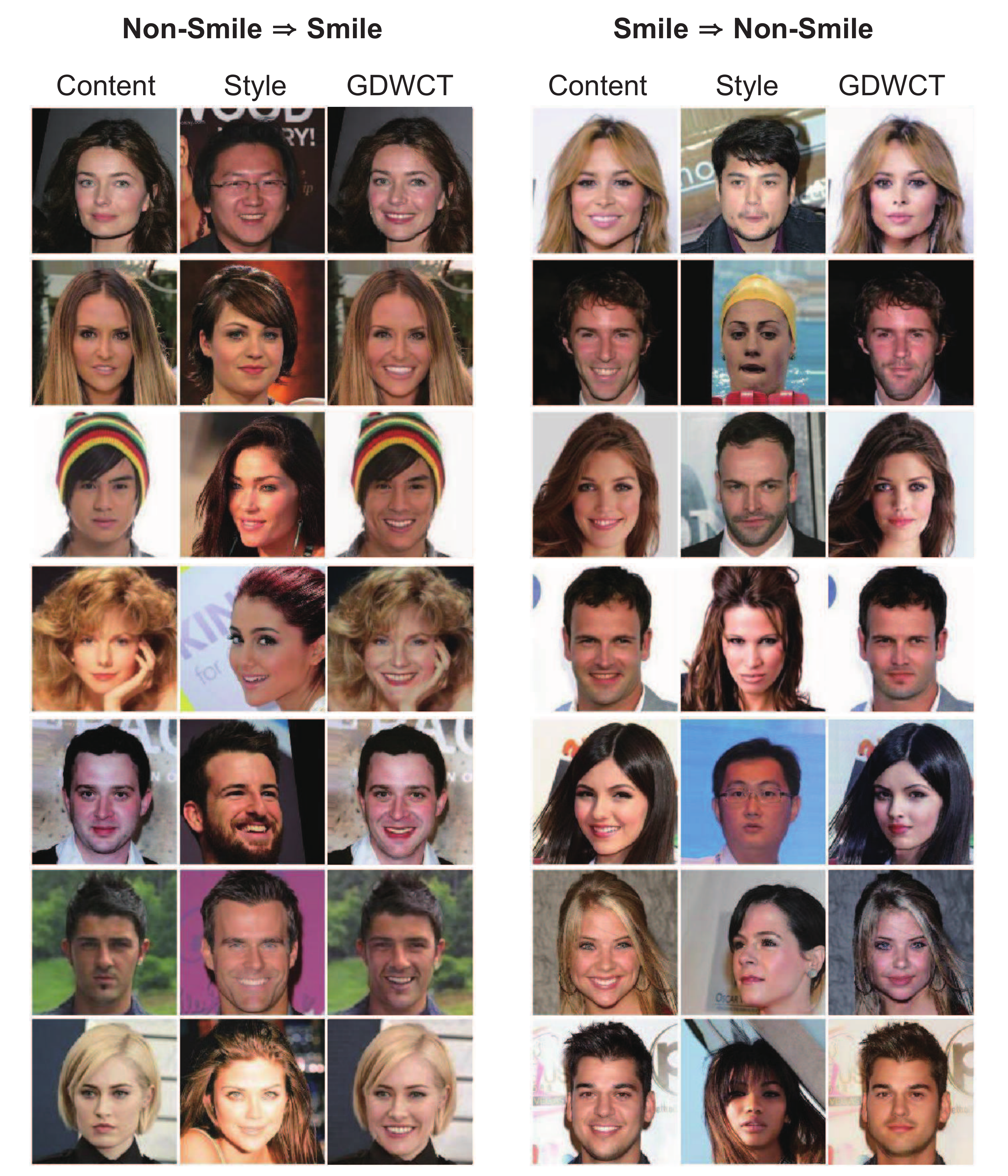}
  \caption{Extra results on CelebA dataset.}
\label{fig:smile}
\end{figure*}

\begin{figure*}[t]
  \includegraphics[width=\linewidth]{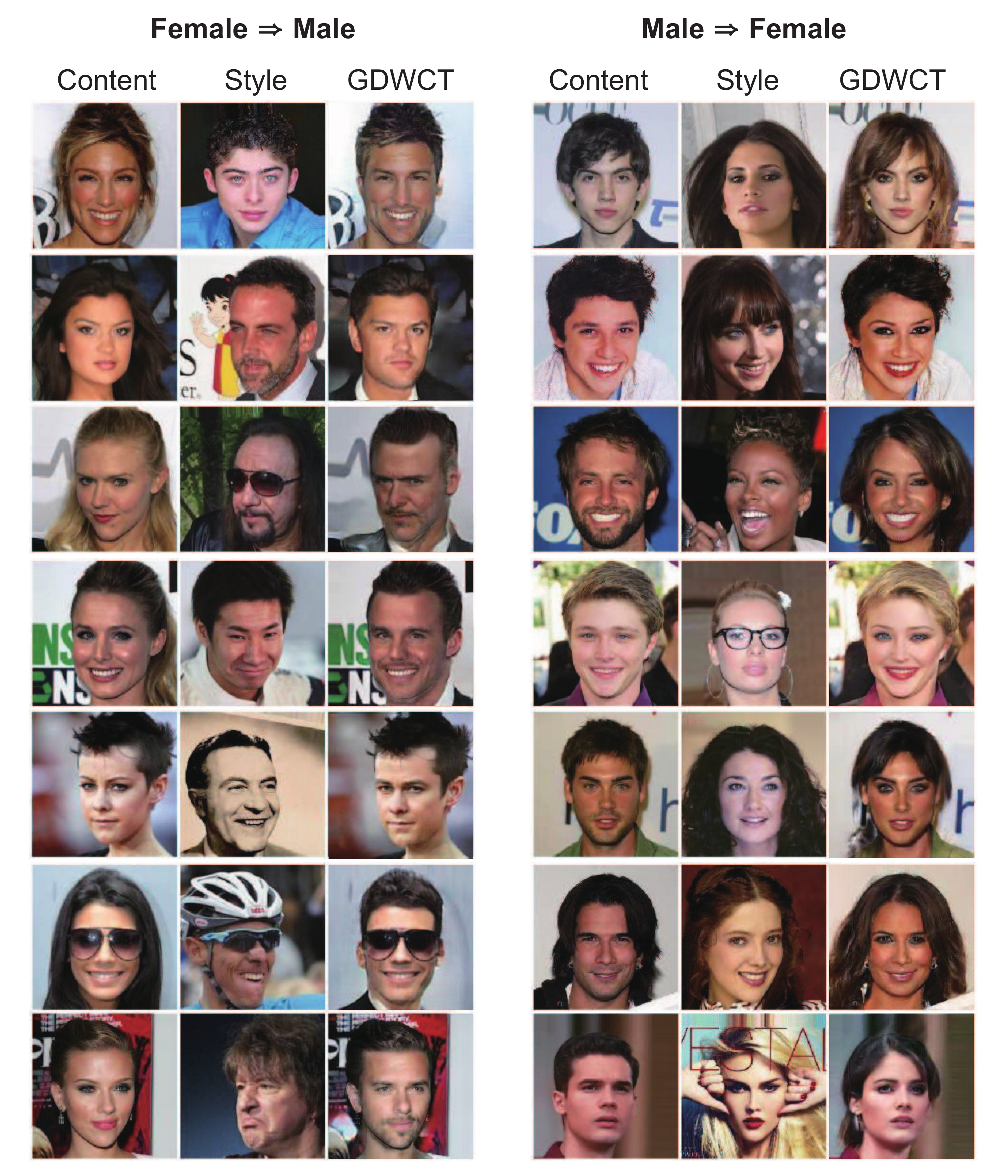}
  \caption{Extra results on CelebA dataset.}
\label{fig:gender}
\end{figure*}

\begin{figure*}[t]
  \includegraphics[width=\linewidth]{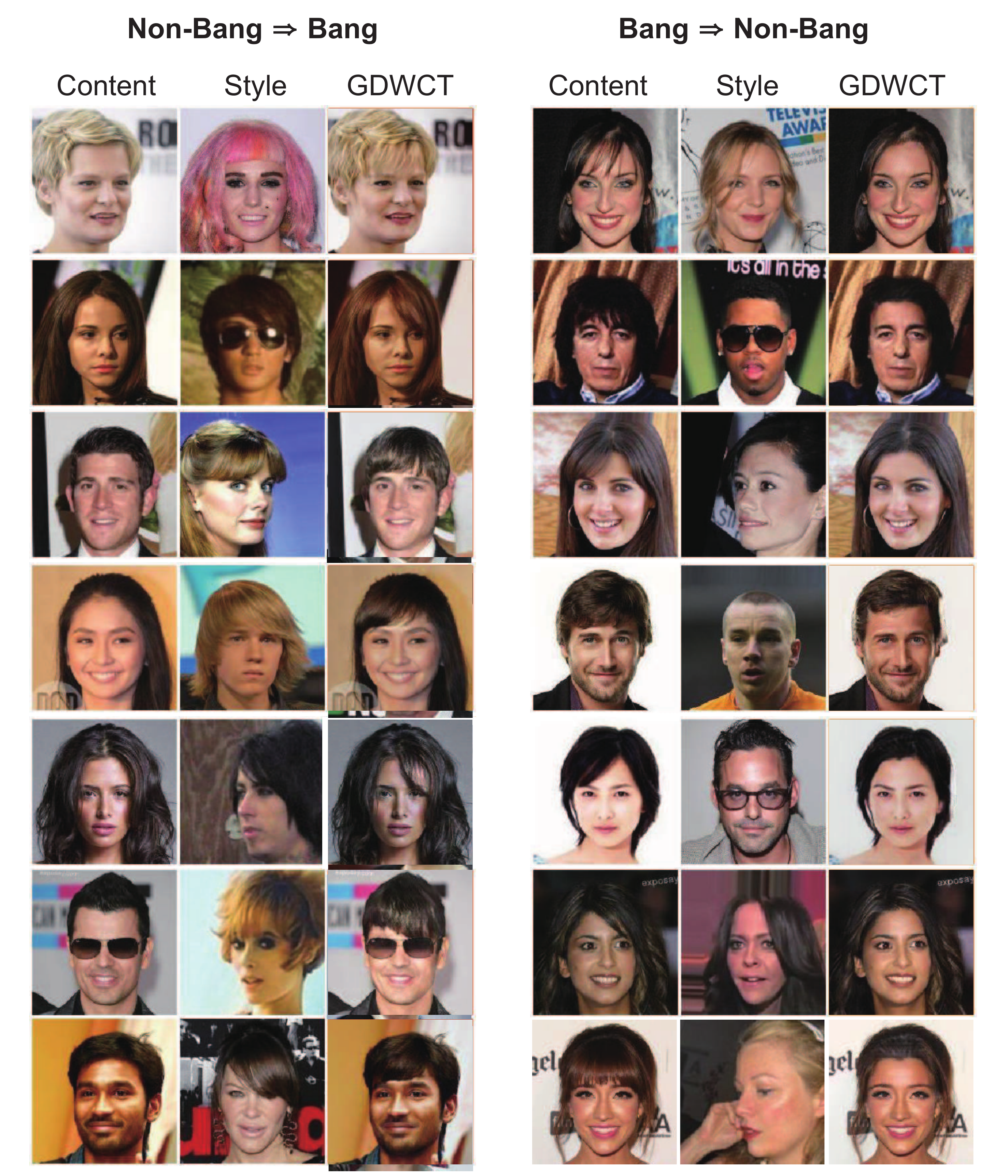}
  \caption{Extra results on CelebA dataset.}
\label{fig:bangs}
\end{figure*}

Finally, we present the extra results of our model in Fig.~\ref{fig:smile},~\ref{fig:gender},~\ref{fig:bangs}. Each translated attribute is written on the top of the macro column. All of the outputs in those figures are generated by the unseen data. Through the results, we verify a superior performance of GDWCT.

\clearpage

\clearpage

\clearpage

\clearpage

\newpage

{\small
\bibliographystyle{ieee_fullname}
\bibliography{egpaper_final}
}

\end{document}